\documentclass{article}

     \PassOptionsToPackage{numbers, compress}{natbib}

    \usepackage[preprint]{neurips_2019}

\usepackage[utf8]{inputenc} 
\usepackage[T1]{fontenc}    
\usepackage{hyperref}       
\usepackage{url}            
\usepackage{booktabs}       
\usepackage{amsfonts}       
\usepackage{nicefrac}       
\usepackage{microtype}      

\usepackage{graphicx}
\usepackage{amssymb}
\usepackage{amsfonts}
\usepackage{amsmath}

\title{Mean field theory for deep dropout networks: \\ digging up gradient backpropagation deeply}

\author{%
  Wei Huang \\
  University of Technology Sydney, Australia \\
  \texttt{Wei.Huang-6student.uts.edu.au} \\
   \AND
   Richard Yi Da Xu \\
   University of Technology Sydney, Australia \\
   \texttt{YiDa.Xu@uts.edu.au} \\
   \AND
   Weitao Du \\
  Northwestern University, USA \\
   \texttt{weitao.du@northwestern.edu} \\
   \And
   Yutian Zeng \\
   Xiamen University, China \\
   \texttt{19020161152850@stu.xmu.edu.cn} \\
   \And
   Yunce Zhao \\
   University of Technology Sydney, Australia \\
   \texttt{Yunce.Zhao@student.uts.edu.au} \\
}

\begin{document}

\maketitle

\begin{abstract}
  In recent years, the mean field theory has been applied to the study of neural networks and has achieved a great deal of success. The theory has been applied to various neural network structures, including CNNs, RNNs, Residual networks, and Batch normalization. Inevitably, recent work has also covered the use of dropout. The mean field theory shows that the existence of depth scales that limit the maximum depth of signal propagation and gradient backpropagation. However, the gradient backpropagation is derived under the {\it gradient independence assumption} that weights used during feed forward are drawn independently from the ones used in backpropagation. This is not how neural networks are trained in a real setting. Instead, the same weights used in a feed-forward step needs to be carried over to its corresponding backpropagation. Using this realistic condition, we perform theoretical computation on linear dropout networks and a series of experiments on dropout networks with different activation functions. Our empirical results show an interesting phenomenon that the length gradients can backpropagate for a {\it single} input and a {\it pair} of inputs are governed by the same depth scale. Besides, we study the relationship between variance and mean of statistical metrics of the gradient and shown an emergence of universality. Finally, we investigate the maximum trainable length for deep dropout networks through a series of experiments using MNIST and CIFAR10 and provide a more precise empirical formula that describes the trainable length than original work.
\end{abstract}

\section{Introduction}

Deep neural networks have achieved exceptional results in a range of fields since its inception \cite{lecun2015deep}. Recent seminal innovations have been proposed to improve the performance of neural networks further. For example, residual networks \cite{he2016deep} and batch normalization \cite{ioffe2015batch}, which were introduced to overcome the gradient vanishing and exploding problem, enabled the trainable length to be very deep. Another technology is the dropout \cite{srivastava2014dropout}, which is a regularization technique for reducing the over-fitting problem. It is also the focus of this paper. In dropout network, units are randomly dropped during training, which can prevent complex co-adaptations \cite{srivastava2014dropout}.

More recently, we have witnessed several signs of progress made using mean field theory \cite{poole2016exponential,schoenholz2016deep,pennington2017resurrecting} in deep learning. The mean field considers networks after random initialization, whose weights and biases were i.i.d. Gaussian distributed, and the width of each layer tends to infinity. As a result of studying signal propagation under mean field theory, an order-to-chaos expressivity phase transition split by a critical line has been found \cite{poole2016exponential}. Later, how parameter initialization may impact the gradient of backpropagation was studied, and the conclusion that the ordered and chaotic phases correspond to regions of vanishing and exploding gradient respectively was shown \cite{schoenholz2016deep}. The results were also equivalently applied to networks with or without dropout. 

The main contribution of the mean field theory for random networks is that it shows the existence of depth scales that limit the maximum depth of signal propagation and gradient backpropagation. Practically, the result is to show a hypothesis that random networks may be trained precisely when information can travel through them. Thus, the depth scales provide bounds on how deep a network may be trained for a specific choice of hyper-parameters \cite{schoenholz2016deep}. This ansatz was tested and verified by practical experiments on MNIST and CIFAR10 dataset with wide width fully-connected networks \cite{schoenholz2016deep}, deep dropout networks \cite{schoenholz2016deep}, and residual networks \cite{yang2017mean}. 

However, the mean field calculation for the gradient is based on the so-called {\it gradient independence assumption}, which states that the weights used during feed forward are drawn independently from the ones used in backpropagation. This is in an effort to make the calculation of gradient feasible regardless of the choice of activation functions. This assumption was later formulated explicitly \cite{yang2017mean} for residual networks and was illustrated in a review \cite{yang2019scaling}. While it enjoys the correct prediction of gradient dynamics in some cases, our experiments show that under the condition in which the weights in feed-forward are carried over to its backpropagation, the length that gradients can backpropagate for a {\it single} and a {\it pair} of inputs are governed by the same depth scale on deep dropout networks instead. 

By further studying the mean and variance of gradient statistics metrics on deep dropout networks, we show an emergence of universality for the relationship between the mean and variance. This universality exists regardless of the choice of hyper-parameters, including dropout rate and activation function. After summarizing the theoretical results about the trainable length of deep dropout networks governed by maximum depth of signal propagation and gradient backpropagation, we perform a series of experiments to investigate it. Empirically, we find a more precise way to describe the maximum trainable length for deep dropout networks, compared with the original results \cite{schoenholz2016deep}.

\section{Related Work}

\begin{figure*}[t!]
\centering
  \centering
  \includegraphics[width=1.0\textwidth]{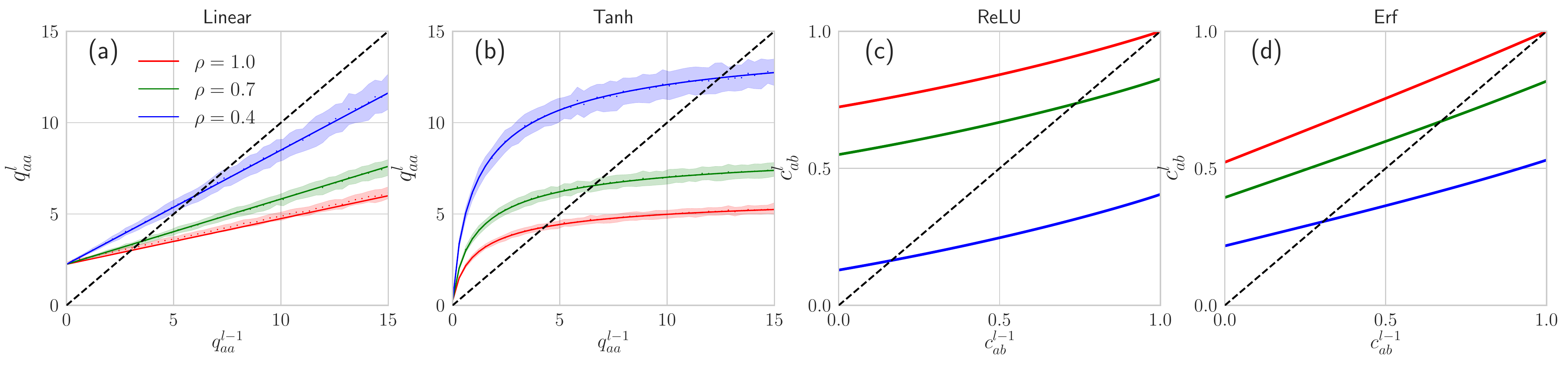}
 \caption{The iterative squared length mapping of Equation (\ref{eq:qlength}) and Equation (\ref{eq:clength}) with different activations and dropout rates. (a) The iterative length map of $q^l_{aa}$ on a Linear network at $\sigma_w = 0.5$ and $\sigma_b = 1.5$. Theoretical predictions (solid lines) match well with network simulations (dots) within a standard error (shadow). The intersection between map and unity line determine its fixed points $q^\ast_{ab}$. Different color correspond to different dropout rates: $\rho = 1$ is red, $\rho = 0.7$ is green, and $\rho = 0.4$ is blue. (b) The iterative length map of $q^l_{aa}$ on a Tanh network at $\sigma_w = 2.5$ and $\sigma_b = 0.5$. (c) The iterative length map of $c^l_{ab}$ on a ReLU network at $\sigma_w = 0.9$ and $\sigma_b = 0.5$. Only intersection of network at $\rho = 1$ (red) is $c^\ast_{ab} = 1$, the others are $c^\ast_{ab} < 1$. (d) The iterative length map of $c^l_{ab}$ on a Erf network at $\sigma_w = 0.9$ and $\sigma_b = 0.5$. Again, $c^\ast_{ab} = 1$ only holds at $\rho = 1$.}
 \label{fig:fixed_point}
\end{figure*}

The mean field theory has been applied to different network architectures, including CNNs \cite{lecun1995convolutional}, RNNs \cite{mikolov2010recurrent}, Residual networks \cite{he2016deep}, Batch normalization \cite{ioffe2015batch}, LSTM \cite{gers1999learning}, and GRUs \cite{chung2014empirical}. These networks have been investigated by \cite{xiao2018dynamical,chen2018dynamical,yang2017mean,yang2019mean,gilboa2019dynamical}, respectively, which form a large family of the mean field theory for deep neural networks. 

Following the mean field theory, \cite{pennington2017resurrecting} studied all singular values of the input-output Jacobian and found a strong connection between {\it dynamical isometry} and fast training speed. Later, the analysis of the spectrum of input-output Jacobian has been developed to provide a detailed analytic understanding \cite{pennington2018emergence} and a nonlinear random matrix theory for deep learning \cite{pennington2017nonlinear}. The study of the spectrum of input-output Jacobian is based on the mean field theory, which will not be addressed in this work since it is trivial to extend the analysis method by \cite{pennington2017resurrecting} to the dropout networks.

In contrast to the mean field theory view to the random networks, \cite{daniely2016toward} studied the relationship between random networks and kernels while \cite{lee2017deep,matthews2018gaussian} adopted another view of Gaussian processes (GPs) in the realm of Bayesian learning. The correspondence between single infinite neural networks and Gaussian process was first observed by \cite{neal1996priors}. Moreover, a study of the dynamics of networks in the infinite width limit, termed as neural tangent kernel, has achieved great success \cite{jacot2018neural,lee2019wide} recently.

Finally, dropout training in deep neural networks can be viewed as approximate Bayesian inference in deep Gaussian processes \cite{gal2016dropout}. Further, dropout can be used in the Neural Network GP \cite{lee2017deep}. While this topic is interesting, we do not include the Bayesian learning of random dropout networks in our work.

\section{Background}

In this section, we review the mean field theory for deep dropout networks. We give the main definitions, setup, and notations, and introduce the results of theory for random networks at initialization, including signal feed-forward and gradient backpropagation, respectively.

\subsection{Feed Forward}

Consider a feed-forward, fully-connected, untrained, and dropout network of depth $L$ with layer width $N$. We denote synaptic weight and bias for the $l$-th layer by $W_{ij}^l$ and $b_{i}^l$; pre-activations and post-activations by $z_i^l$ and $y_i^l$ respectively. Finally, we take the input to be $y^0_i = x_i$ and the dropout keep rate to be $\rho$. The information propagation in this network is governed by,
  \begin{equation} \label{eq:propagation}
    z^l_i =  \frac{1}{\rho} \sum_{j} W_{ij}^l p^l_j y_j^{l-1} + b_i^l,~~~y_i^l  = \phi(z_i^l), 
  \end{equation}
where $\phi: \mathbb{R} \rightarrow  \mathbb{R}$  is the activation function and $p \sim$ Bernoulli($\rho$). We adopt the mean field theory assumption \cite{poole2016exponential,schoenholz2016deep}, where $W^l_{ij} \sim  \mathcal{N} (0, \frac{\sigma^2_w}{N} ) $, $ b_i^l \sim \mathcal{N} (0,\sigma^2_b)$, and the width $N$ tends to infinite. Since the weights and biases are randomly distributed, these equations define a probability distribution on the pre-activations over an ensemble of untrained neural networks. Under the mean field approximation, $z^l_i$ can be replaced by a Gaussian distribution with zero mean.

Consider a {\it single} input $x_{i;a}$, where the subscript $a$ refers to the index of input. We define the length quantities $q_{aa}^l = \frac{1}{N} \sum_{i=1}^N (z^l_{i;a})^2$, which is the mean squared pre-activations. According to the mean field approximation, the length quantity is described by the recursion relation, 
 \begin{equation}\label{eq:qlength}
 \begin{aligned}
    q_{aa}^l & = \frac{\sigma_w^2}{\rho} \int \mathcal{D}z \phi^2(\sqrt{q_{aa}^{l-1}}z) + \sigma^2_b, \\
 \end{aligned} 
 \end{equation}
where $ \int \mathcal{D}z = \frac{1}{\sqrt{2\pi}} \int dz e^{-\frac{1}{2}z^2} $ is the measure for a normal distribution. This equation describes how a single input evolves through a random neural network. To study the property of evolution, we investigate the fixed point at $q^{\ast}_{aa} \equiv \lim_{l \rightarrow \infty} q^l_{aa}$. One way to estimate the fixed point is to plot Equation (\ref{eq:qlength}) with the unity line, and the intersection is the fixed point. We show the result for Equation (\ref{eq:qlength}) with Linear dropout network and Tanh dropout network in Figure \ref{fig:fixed_point}(a)(b). Note that the smaller the dropout rate $\rho$, the larger the fixed point value $q^\ast_{aa}$.

The propagation of a {\it pair} of inputs $x_{i;a}$ and $x_{i;b}$, where the subscript $a$ and $b$ refer to different inputs, can be studied by looking at the correlation between the two inputs after $l$ layers. We definite this correlation quantity as $q^l_{ab} = \frac{1}{N} \sum_{i=1}^N (z^l_{i;a} z^l_{i;b})$. Similarly, the correlation $q^l_{ab}$ will be given by the recurrence relation,
\begin{equation}\label{eq:c1length}
q_{ab}^l  = \sigma_w^2 \int \mathcal{D}z_1 \mathcal{D}z_2 \phi(u_1)\phi(u_2) + \sigma^2_b,
\end{equation}
where $u_1 = \sqrt{q^{l-1}_{aa}} z_1$ and $u_2 = \sqrt{q^{l-1}_{bb}} (c^{l-1}_{ab}z_1+\sqrt{1-(c^{l-1}_{ab})^2}z_2)$, with 
\begin{equation}\label{eq:clength}
 c^l_{ab} = q^l_{ab}/\sqrt{q^l_{aa}q^l_{bb}}.
\end{equation}
This equation also have a fixed point at $c^{\ast}_{ab} \equiv \lim_{l \rightarrow \infty} c^l_{ab}$. It is known that $c^\ast_{ab} = 1$ when $\rho=1$, while $c^\ast_{ab} < 1$ when $\rho<1$ \cite{schoenholz2016deep}. We show the result of Equation (\ref{eq:clength}) on the ReLU and Erf dropout networks in Figure \ref{fig:fixed_point}(c)(d), which demonstrate the main conclusion about fixed-point without ($\rho=1$) and with ($\rho<1$) dropout. 

The main contribution of mean field theory for the fully-connected networks without dropout ($\rho=1$) is that it presents a phase diagram, which is determined by a crucial quantity,
\begin{equation}\label{eq:chi1}
  \chi_1 = \frac{\partial c^l_{ab}}{\partial c^{l-1}_{ab}} = \sigma_w^2 \int \mathcal{D}z [\phi'(\sqrt{q^\ast}z)]^2.
\end{equation}
This quantity was firstly introduce by \cite{poole2016exponential} to determine whether or not the $c^\ast_{ab}= 1$ is an attractive fixed point. When $\chi_1 > 1$, the fixed point is unstable. Conversely, when $\chi_1 < 1$, the fixed point is stable. Thus, the critical line $\chi_1=1$ separates two phases. One is the chaotic phase ($\chi_1 > 1$), where a pair of inputs end up asymptotically decorrelated, and the other is the ordered phase ($\chi_1 < 1$), in which a pair of inputs end up asymptotically correlated.

We give a comment on the difference between $q^l_{aa}$ and $c^l_{ab}$ here. The random networks in the infinite width limit can be viewed as the Gaussian processes, where $q^l_{aa}$ and $c^l_{ab}$ are the diagonal and non-diagonal elements of the compositional kernel \cite{lee2017deep}, respectively. Intuitively, the non-diagonal element of the kernel measures the correlation between different data points while the diagonal component measures the information of one input itself. 

The study of information propagation shows the existence of a depth-scales $\xi_2$, which represent the length of propagation of the following qualities:
 \begin{equation}\label{eq:pro}
   |c_{ab}^l - c_{ab}^\ast|  \sim e^{-l/\xi_2}.
 \end{equation}
where  $\xi_2 = |1/\log \chi_2|$, with $\chi_2 =  \sigma_w^2  \int \mathcal{D}z_1 \mathcal{D}z_2 \phi'(u^\ast_1) \phi'(u^\ast_2)$, where $u^\ast_1 = \sqrt{q^{\ast}_{aa}} z_1$ and $u^\ast_2 = \sqrt{q^{\ast}_{bb}} (c^{\ast}_{ab}z_1+\sqrt{1-(c^{\ast}_{ab})^2}z_2)$. Intuitively, the depth-scales $\xi_2$ measures how far can correlation between two different inputs survives through the network. 

\subsection{Back Propagation} 

There is a duality between the forward propagation of signals and the backpropagation of gradients. Given a loss $E$, we have
  \begin{equation}  \label{delta}
    \frac{\partial E}{\partial W^{l}_{ij}} = \frac{ p^l_j}{\rho}  \phi(z^{l-1}_j) \delta^l_i, ~~~ \delta^l_i = \phi'(z^l_i)\frac{ p^{l+1}_i}{\rho} \sum_j \delta^{l+1}_j   W^{l+1}_{ji},  
  \end{equation}
where $\delta^l_i =\frac{\partial E}{\partial z^l_i} $. We define the metric of gradient for both a {\it single} input and a {\it pair} of inputs cases:
 \begin{equation}
   g^l_{aa} \equiv \frac{1}{N^2} \sum_{ij} (\frac{\partial E_a}{\partial W^l_{ij}})^2,~~~  g^l_{ab} \equiv \bigg | \frac{1}{N^2}\sum_{ij} \frac{\partial E_a}{W^l_{ij}} \frac{\partial E_b}{W^l_{ij}} \bigg|.
 \end{equation}

Within mean field theory, the scale of fluctuations of the gradient of weights in a layer will be proportional to $\tilde{q}^l_{aa} \equiv \mathbb{E}\left[ \delta^l_{i;a} \delta^l_{i;a} \right] $, which can be written as, $g^l_{aa} \propto \tilde{q}^l_{aa}$ \cite{schoenholz2016deep}. On the other hand, the correlation between gradients of a pair of inputs will be proportional to $ \tilde{q}^l_{ab} \equiv \mathbb{E}\left[ \delta^l_{i;a} \delta^l_{i;b} \right]$, namely,  $g^l_{ab} \propto \tilde{q}^l_{ab}$.

In order to work out the recurrence relation for $\tilde{q}^l_{aa}$ and  $\tilde{q}^l_{ab}$, an approximation was made \cite{schoenholz2016deep}, named {\it gradient independence assumption}, that the weights used during forward propagation are drawn independently from the weights used in backpropagation. In this way, the term $\phi'(z_i^l)$, $\delta_j^{l+1}$ and $W_{ji}^{l+1}$ in Equation (\ref{delta}) can be addressed independently. Then, the recurrence behavior of $\tilde{q}^l_{aa}$ and $\tilde{q}^l_{ab}$ are achieved,
  \begin{equation} \label{gradient}
   \tilde{q}^l_{aa} = \tilde{q}^{l+1}_{aa} \chi_1,~~~\tilde{q}^l_{ab} = \tilde{q}^{l+1}_{ab} \chi_2.
  \end{equation}
where we redefine the quantity $\chi_1$ for the dropout networks,
\begin{equation}\label{eq:chi1}
  \chi_1 = \frac{\sigma_w^2}{\rho} \int \mathcal{D}z [\phi'(\sqrt{q^\ast}z)]^2.
\end{equation}
Equation (\ref{gradient}) has an exponential solution with,
\begin{equation} 
   \tilde{q}^l_{aa} = \tilde{q}^{L}_{aa} e^{-(L-l)/\xi_1},~~~\tilde{q}^l_{ab} = \tilde{q}^{L}_{ab} e^{-(L-l)/\xi_2}.
  \end{equation}
Similar to the signal propagation, gradient backpropagation can limit the trainable length in the way of gradient vanishing or gradient exploding, which is measured by the depth-scales $\xi_1$ and $\xi_2$.

\begin{figure*}[t!]
\centering
  \centering
  \includegraphics[width=1.0\textwidth]{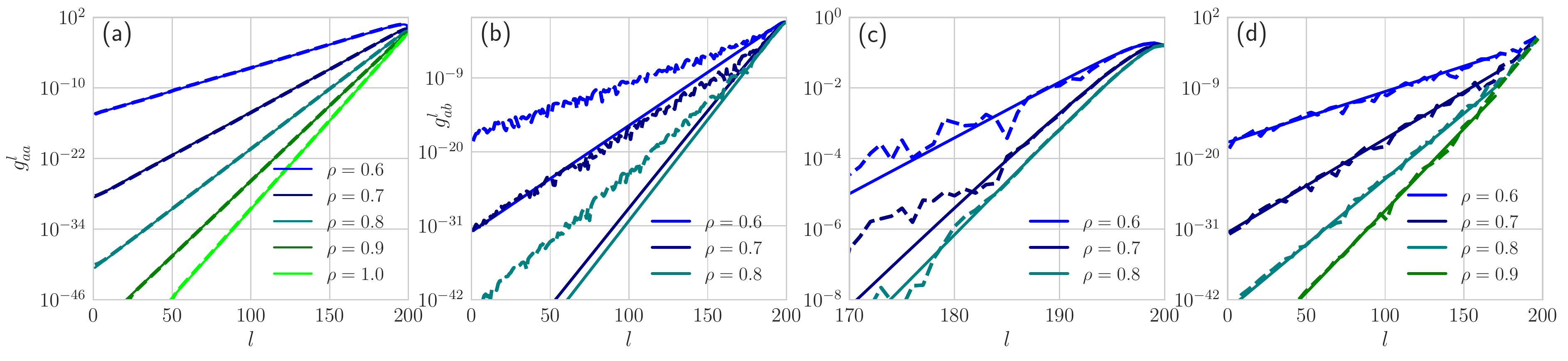}
 \caption{Theoretical calculations versus network simulations for metric of gradient. (a) $g^l_{aa}$ as a function of layer $l$, for a 200 layers random linear network with $\sigma_w^2 = 0.5$ and $\sigma_b^2 =0.1$. Excellent agreement is observed between empirical simulations of networks of width 1000 (dashed lines) and theoretical calculations (solid lines). (b) $g^l_{ab}$ as a function of layer $l$. Theoretical calculations (solid lines) fail to predict empirical simulations (dashed lines). (c) $g^l_{ab}$ as a function of layer $l$ in the range of length $l=170-200$. Theoretical calculations (solid lines) can predict empirical simulations (dashed lines) in the few last layers. (d) $g^l_{ab}$ as a function of layer $l$. The solid lines are $g^l_{ab} \propto \chi_1^{L-l} $ for different $\rho$. Theoretical calculations failed to predict empirical simulations (dashed lines).}
 \label{fig:grad}
\end{figure*}

\begin{figure*}[t!]
\centering
  \centering
  \includegraphics[width=0.8\textwidth]{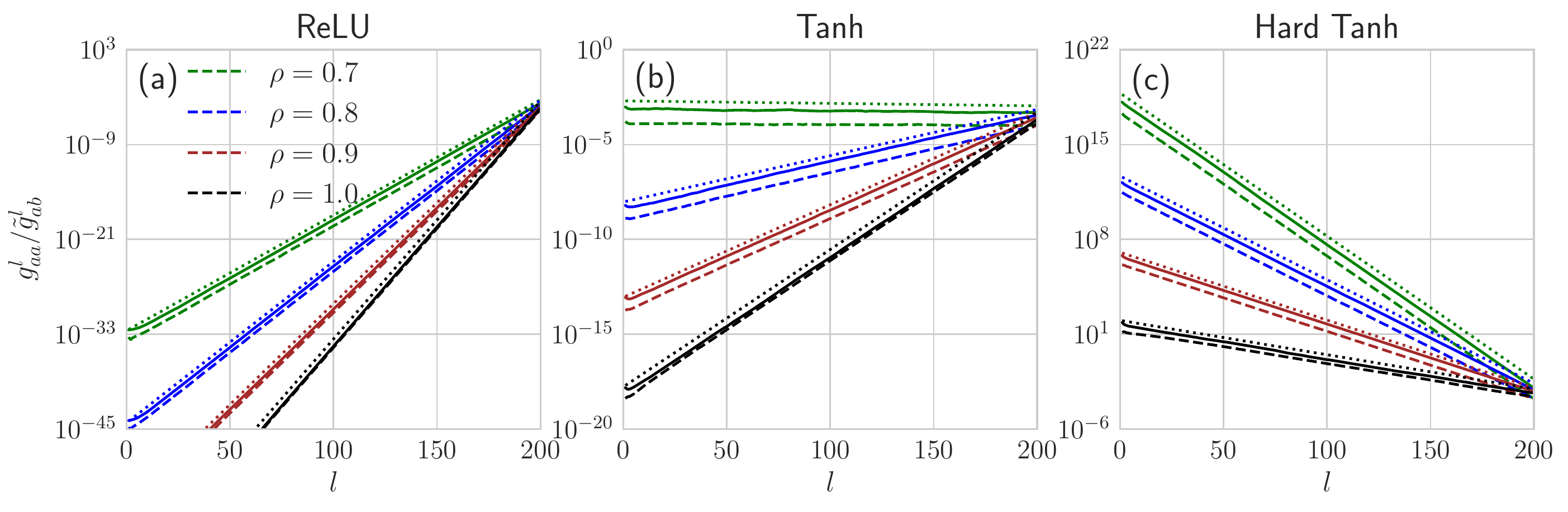}
 \caption{The metric of gradient with one and two different inputs, $g^l_{aa}$ (solid lines), $\tilde{g}^l_{ab}$ (dashed lines), and $g^l \propto \chi_1^{L-l}$ (dotted lines) as a function of layer $l$ with different activation. (a) ReLU network with $\sigma^2_w =1.0$ and $\sigma^2_b = 0.1$. (b) Tanh network with $\sigma^2_w =1.4$ and $\sigma^2_b = 0.1$. (c) Hard Tanh network with $\sigma^2_w =1.4$ and $\sigma^2_b = 0.1$.  Excellent agreement is observed between empirical simulations of $g^l_{aa}$, $\tilde{g}^l_{ab}$, and formula $g^l = g^{l+1} \chi_1 $.}
 \label{fig:gradsim}
\end{figure*}

\section{Gradient Backpropagation}

In this section, we first calculate the metrics of gradient $g_{aa}$ and $g_{ab}$ theoretically without the gradient independence assumption on linear dropout networks. We then conduct a series experiment for metrics of gradient on deep dropout networks, including non-linear cases. Finally, we show an emergence of a universal relationship between mean and variance of metrics of the gradient. 

\subsection{Breaking the gradient independence assumption}

We follow the fact that weights used in a feed-forward are carried over to its back-propagation. We first provide a theoretical treatment to the linear networks in which we assume the output is the last layer of network $y^L_i =  z^L_i$ without soft-max. The labels of data are set to be zeros, and the loss is the mean squared loss. 

For space reason, we omit details of the calculation and present the primary analysis and final results here. The main problem is that we should expand $\delta^{l+1}_j$ when calculating $\delta^l_i$ in Equation (\ref{delta}), since $\delta^{l+1}_j$ can correlate with $W_{ji}^{l+1}$ without the gradient independence assumption. Using $g^l_{aa}$ as an example, we perform:
\begin{enumerate}
\item Starting from the last layer $L$, we compute $\delta^L_{i,a} = \frac{\partial E_a}{ \partial z^L_{i,a}} = 2 z^L_{i,a}$ and use this result to compute $g^L_{aa} = \mathbb{E}\left[ (\frac{ p^L_{j,a}}{\rho}  z^{L-1}_{j,a} \delta^L_{i,a})^2 \right] $.
\item Then we compute $g^{L-1}_{aa} = \mathbb{E}\left[ (\frac{ p^{L-1}_{j,a}}{\rho}  z^{L-2}_{j,a} \delta^{L-1}_{i,a})^2 \right ]$ with the result of $\delta_{i,a}^{L-1} = \frac{\partial E_{a}}{\partial z_{i,a}^{L}}\frac{\partial z_{i,a}^{L}}{\partial z_{i,a}^{L-1}}=\sum_{j}2z_{j,a}^{L}\frac{p_{i,a}^{L}}{\rho}W_{ji}^{L}$ and $z^L_i =  \frac{1}{\rho} \sum_{j} W_{ij}^L p^l_j z_j^{L-1} + b_i^L$.
\item  By parity of reasoning, we obtained the results for the penultimate layer $g_{aa}^{L-2}$. The correlation between terms that contain $W_{ij}^L$ and $W_{ij}^{L-1}$ are considered.
\item  As the index of the layer decreases, the amount of calculation becomes larger and larger. Thus we use the induction method to achieve the results for left layers.
\end{enumerate}
We use the same approach to derive the result for $g^l_{ab}$. As a result, we have,
  \begin{equation} \label{eq:gaa}
   \begin{aligned}
   g^l_{aa} & = 4 (\frac{q^\ast_{aa}}{\rho})^2 (\frac{\sigma^2_w}{\rho})^{L-l} [\rho+\sum_{j=1}^{L-l}  (\frac{\sigma^2_w}{\rho})^{j}], \\
   g^l_{ab} & = 4 (q^\ast_{ab})^2 (\sigma^2_w)^{L-l} [1 +\sum_{j=1}^{L-l}  (\frac{\sigma^2_w}{\rho^2})^{j}].
  \end{aligned}
  \end{equation}

\begin{figure*}[t!]
\centering
  \centering
  \includegraphics[width=0.8\textwidth]{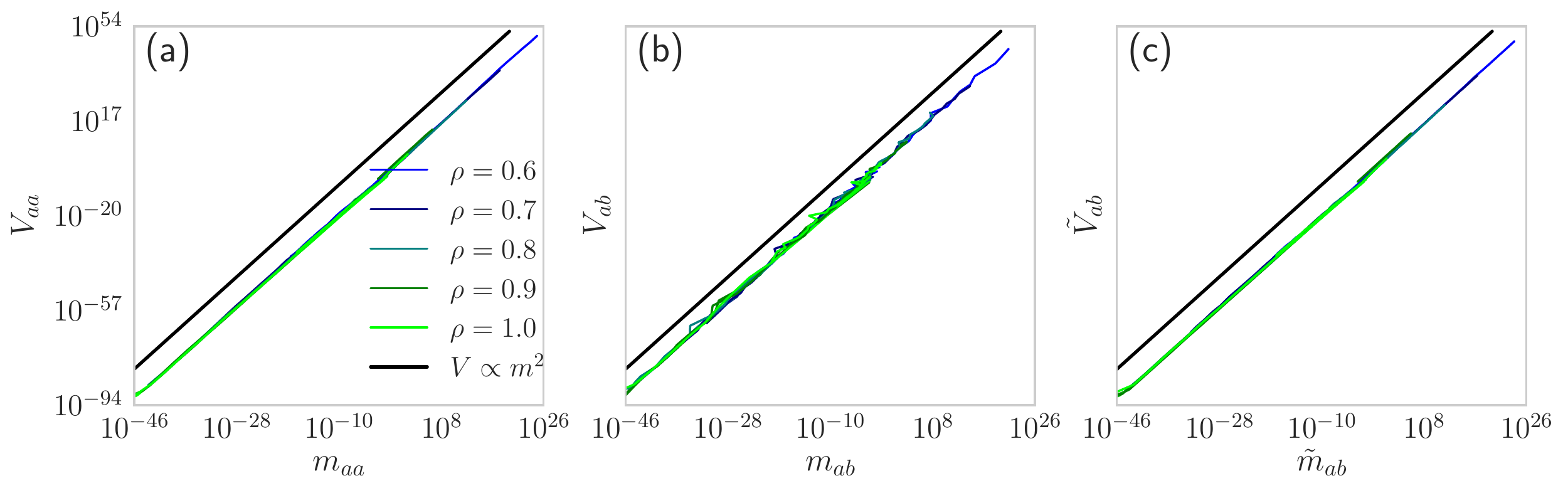}
 \caption{Universal relationship between variance and mean of $g^l_{aa}$, $g^l_{ab}$, and $\tilde{g}^l_{ab}$, on the 200 layers and width $N=500$ random dropout networks. Different color represents a different dropout rate. The black line is the function of $V \propto m^2$. (a) $V^l_{aa}$ as a function $m^l_{aa}$. (b) $V^l_{ab}$ as a function of $m^l_{ab}$. (c) $\tilde{V}^l_{ab}$ as a function of $\tilde{m}^l_{ab}$. All the curves regarding different activations collapse to a line, and the power coefficient of all curves is consistent with 2.}
 \label{fig:allvm}
\end{figure*}

\begin{figure*}[t!]
\centering
  \centering
  \includegraphics[width=0.8\textwidth]{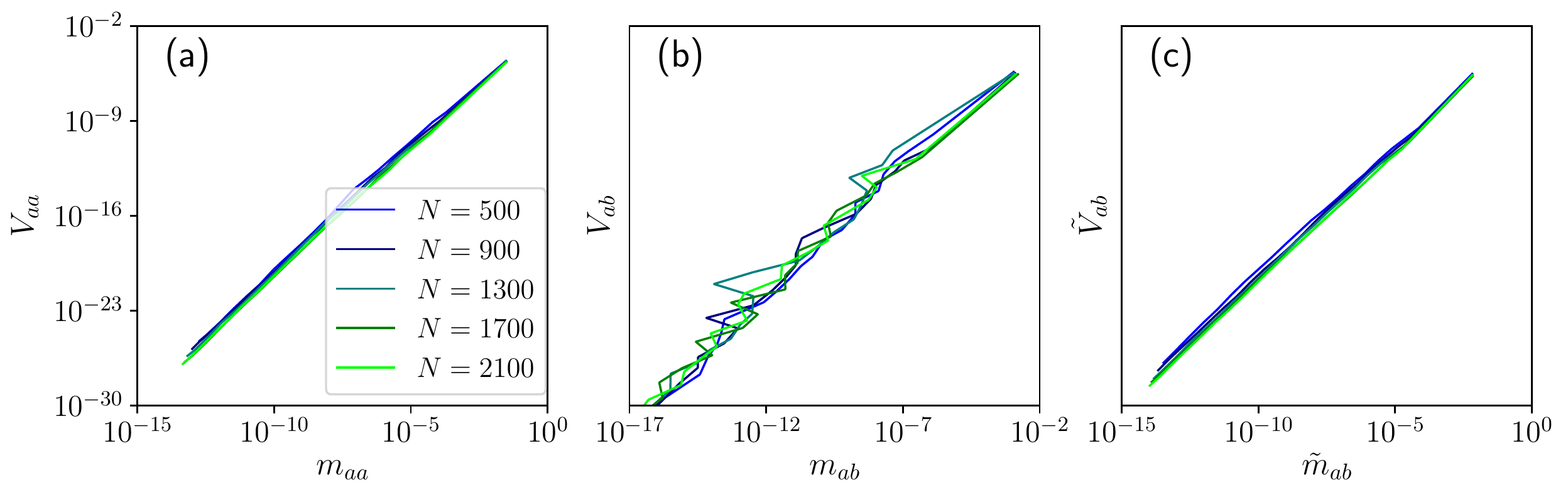}
 \caption{Universal relationship between variance and mean of $g^l_{aa}$, $g^l_{ab}$, and $\tilde{g}^l_{ab}$, on the 200 layers, Tanh random dropout networks with $\rho = 0.9$. All the curves regarding different width collapse to a line. Different color represents a different network width. (a) $V^l_{aa}$ as a function $m^l_{aa}$. (b) $V^l_{ab}$ as a function of $m^l_{ab}$. (c) $\tilde{V}^l_{ab}$ as a function of $\tilde{m}^l_{ab}$.}
 \label{fig:Nvm}
\end{figure*}

By analyzing the first formula of Equation (\ref{eq:gaa}), we find that $g^l_{aa} = g^{l+1}_{aa} \chi_1 $. This can be better observed by dividing the expression related to layer $l$ into two factors: one is $(\frac{\sigma^2_w}{\rho})^{L-l}$, and the other is $\sum_{j=1}^{L-l} (\frac{\sigma^2_w}{\rho})^{j}$. The first factor accounts for $g^l_{aa} = g^{l+1}_{aa} \chi_1 $, where $\chi_1 = \frac{\sigma^2_w}{\rho}$ for linear dropout networks. And second factor will be stable after several layers starting from the last layer $L$ due to $\sigma^2_w < \rho$. We show an excellent match between the theoretical calculation above with simulation using networks with width $N=500$ and layer $L=200$ over 100 different instantiations of the network in Figure \ref{fig:grad}(a). 

Despite the successful prediction of theoretical calculation for $g^l_{aa}$, our theoretical results for $g^l_{ab}$ only hold on the case of $\rho=1$ while fail to predict the experimental behavior except for last few layers when $\rho < 1$, as shown in Figure \ref{fig:grad}(b)(c). After a few layers from $L$, the variances began to increase dramatically as shown in Figure \ref{fig:grad}(c). We noticed that unlike the case of computing $q^l_{ab}$, using $\chi_2$ is prohibitive for computing $g^l_{ab}$. On the other hand, we try a function regarding $\chi_1$ to fit $g^l_{ab}$, and find an interesting observations that $\chi_1$ is a much more compatible term for $g^l_{ab}$, i.e, $g^l_{ab} =  g^{l+1}_{ab}\chi_1$. This is demonstrated in Figure \ref{fig:grad}(d). 

The incompatible phenomenon between theoretical calculation and experimental results for $g^l_{ab}$ begins with the emergence of variance, as shown in Figure \ref{fig:grad}(c). One possible explanation is that the emergence of variance is caused by limited network length. Thus, we can reduce this variance by increasing network length only. To check if this explanation works, we further investigate the relationship between variance and mean of $g^l_{ab}$ with different network widths $N$. The answer is that  $g^l_{ab} =  g^{l+1}_{ab}\chi_1$ holds regardless of the finite width. We will demonstrate it in the next section.

After studying the gradient behavior at the linear networks, we conduct a series of experiments on the nonlinear case since the theoretical formulation for nonlinear activation or with the soft-max layer is intractable. We firstly use $g^l_{ab}$ as the metric of gradient and find it has a huge variance when $\rho<1$. This is because the element of the gradient matrix with a pair of inputs can be either negative or positive. To find a metric with low variance, we consider the metric $\tilde{g}_{ab}^l \equiv \frac{1}{N^2}\sum_{ij} |\frac{\partial E_a}{W^l_{ij}} \frac{\partial E_b}{W^l_{ij}}| $ whose elements are all positive. Besides, it is the $\ell_1$ norm of the gradient matrix. 

We plot $g^l_{aa}$ and $\tilde{g}_{ab}^l$ as a function of $l$ in Figure \ref{fig:gradsim}. Interestingly, our simulations show that both $g^l_{ab}$ and $\tilde{g}_{ab}$ are governed by $\chi_1$ in a range of activations. Thus we make a conjecture that the relation,
  \begin{equation}\label{eq:gchi} 
   g^l_{aa} = g^{l+1}_{aa} \chi_1,~~~g^l_{ab} = g^{l+1}_{ab} \chi_1,
  \end{equation}
holds on deep dropout networks.

\begin{table*}[!htbp]
		\caption{Summary of depth-scale for theoretical results i.e. signal propagation and gradient backpropagation, and empirical results under different condition or assumption.}
		\label{tab:depth}
		\begin{center}
			\begin{tabular}{|c |c c| c c| c|}
			\hline
			 Summary    & \multicolumn{2}{c|}{feed-forward propagation}  & \multicolumn{2}{c|}{gradient backpropagation}  & empirical results \\
			\hline 
			metric  & ~~~~$q_{aa}$ & $q_{ab}$  &~~~~~ $g_{aa}$ & $g_{ab}$   &  \\
         \hline
		  realistic condition  (our work)      &  ~~~ - & $\xi_2$    &~~~~~ $\xi_1$ & $\xi_1$   & $\min \{12\xi_1, 12\xi_2 \}$ \\
        independent assumption \cite{schoenholz2016deep}   &  ~~~~- & $\xi_2$    &~~~~~ $\xi_1$ & $\xi_2$ & $6 \xi_2$ \\ 
			\hline
			\end{tabular}
		\end{center}
	\end{table*}	

\subsection{Emergence of Universality}

\begin{figure*}[t!]
\centering
  \centering
  \includegraphics[width=0.9\textwidth]{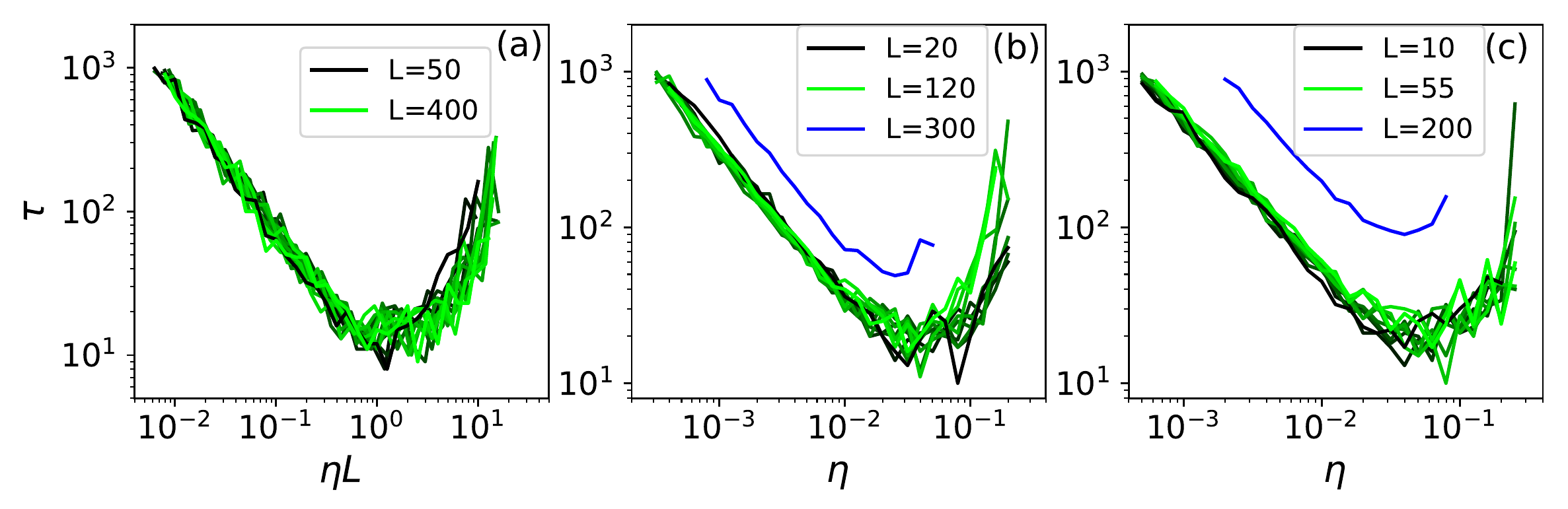}
 \caption{The number of steps $\tau$ to reach test accuracy $p \approx 0.25$ as a function learning rate $\eta$. (a) Network without dropout, colors reflect different network depth $L$ from 50 (black) to 400 (green). They all collapse to a single universal curve when the learning rate $\eta$ is re-scaled by $L$. (b) Network with dropout $\rho =0.99$, colors reflect different network depth $L$ from 20 (black) to 120 (green), additional $L=300$ is colored blue for comparison. Curves with $L \le 120$ collapse to a universal curve without any re-scale. (c) Network with dropout $\rho =0.98$, colors reflect different network depth $L$ from 10 (black) to 55 (green), additional $L=200$ is colored blue for comparison. Curves with $L \le 55$ collapse to a universal curve without any re-scale.}
 \label{fig:speed}
\end{figure*}

We have studied three statistical metrics of the gradient, i.e. $g_{aa}$, $g_{ab}$, and $\tilde{g}_{ab}$ using their mean value. Inevitably, the variance of these metrics can give us essential information about the gradient. To do this, we performed a series of experiments to obtain the mean and variance of $g_{aa}$, $g_{ab}$ and $\tilde{g}_{ab}$ with different activation and different network width $N$.

First, we show the relationship between variance and mean of the metric of gradient with different activations, including Linear, ReLU, Tanh, and Hard Tanh. We denote the mean of $g_{aa}$, $g_{ab}$ and $\tilde{g}_{ab}$ as $m_{aa}^l$, $m_{ab}^l$, and $\tilde{m}^l_{ab}$, while naming the variance as $V_{aa}^l$, $V_{ab}^l$, and $\tilde{V}^l_{ab}$ respectively. We show the variance as a function of mean in Figure \ref{fig:allvm}, and find the emergence of universality between the variance and mean regardless of dropout rate and choice of activation for $g_{aa}$, $g_{ab}$, and  $\tilde{g}^l_{ab}$. 

The plot of variance as a function of mean shows a power-law between them since it is like a straight line in the log-log plot. To estimate the power, we use a simple equation $V \propto m^2$ to compare with the experiment results. Surprisingly, all three curves are consistent with $V \propto m^2$. Thus we make a conjecture that the universal power coefficient between the variance and mean is 2.

Then, we investigate the relationship between variance and mean with different network width $N$ and show the results in Figure \ref{fig:Nvm}. This time, we perform experiments on the $\rho = 0.9$ Tanh networks with different network width $N$. Again, the relationship between variance and mean satisfies universality, which means the Equation (\ref{eq:gchi}) does not depend on the network width of $N$. 

We want to point out that we have performed the same investigation on $q^l_{aa}$ and $c^l_{ab}$. However, we did not observe a similar universal relationship between variance and mean of $q^l_{aa}$ and $c^l_{ab}$. This may occur due to the different behavior of $q_{aa}^l$ ($q_{ab}^l$) and $g_{aa}^l$ ($g_{ab}^l$). As Equation (\ref{eq:pro}) shows, the mean of $c^l_{ab}$ will converge to a fixed point after several layers, which means that the mean of $c^l_{ab}$ will be stable in deeper layers. So, we won’t expect a universal relation between the mean and the variance in this case.

In summary, we have tried all the parameter freedom that we can tune, the universal power coefficient between the variance and mean remains the same. We conclude that once the topological structure of the neural network is set, the power coefficient is universal.

\begin{figure*}[t!]
\centering
  \centering
  \includegraphics[width=1.05\textwidth]{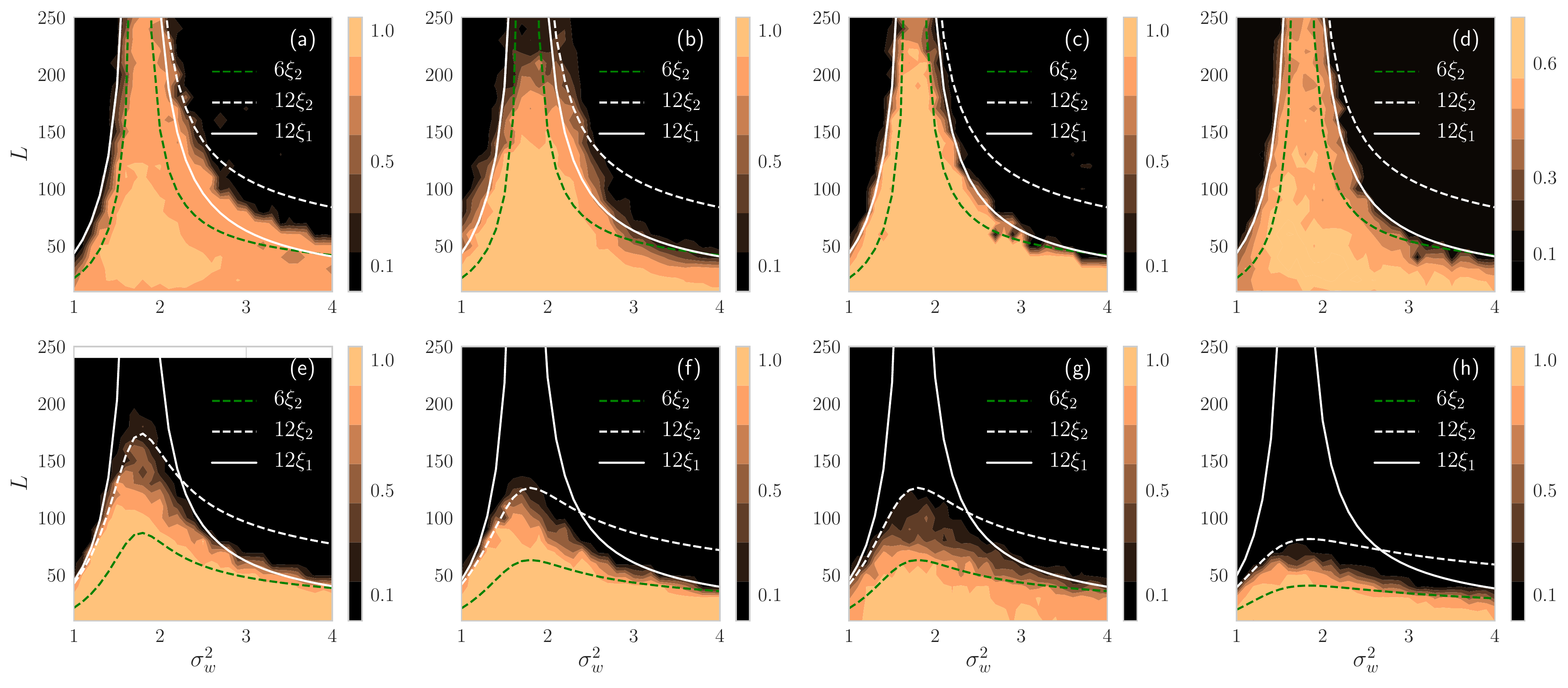}
 \caption{The training accuracy for neural networks as a function of the depth $L$ and initial weight variance $\sigma^2_w$ from a high accuracy (bright yellow) to low accuracy (black). Comparison is made by plotting $12\xi_1$ (white solid line), $6\xi_2$ (green dashed line), and $12\xi_2$ (white dashed line). (a) 2000 training steps of $\rho=1$ network with Gaussian weights on the MNIST using SGD. (b) 1000 training steps of $\rho=1$ network with Gaussian weights on the MNIST using RMSProp. (c) 2000 training steps of $\rho=1$ network with Orthogonal weights on the MNIST. (d) 3000 training steps of $\rho=1$ network with Orthogonal weights on CIFAR10. (e) 3000 training steps of $\rho=0.99$ network with Orthogonal weights on the MNIST.  (f) 3000 training steps of $\rho=0.98$ network with Orthogonal weights on the MNIST using SGD. (g) 10000 training steps of $\rho=0.98$ network with Gaussian weights on the MNIST. (h) 3000 training steps of $\rho=0.95$ network with Orthogonal weights on the MNIST using SGD.}
 \label{fig:heatmap}
\end{figure*}

\section{Experiments}

According to the theoretical results, during feed-forward, we expect that length-scale $\xi_2$ controls the propagation of $c^l_{ab}$, while $\xi_1$ measures the number of layers that gradient metrics $g^l_{aa}$ and $g^l_{ab}$ can survive during backpropagation. However, \cite{schoenholz2016deep} claimed that both networks with or without dropout networks have a limited trainable length, which is governed by the depth-scale $\xi_2$. As our experimental results show, which be demonstrated later, this statement is not exactly right. To summarize, we present the comparison for the length-scale between \cite{schoenholz2016deep} and our work in Table \ref{tab:depth}.

\subsection{Training speed}

Before investigating this problem, we study the relationship between training speed and choice of hyper-parameters. We confine the hyper-parameters at the critical line $\chi_1 = 1$ for the network with and without dropout and train networks of a range of length with width $N=400$ for $10^3$ steps with a batch size of $10^3$ on the standard CIFAR10 dataset. Strictly speaking, $\chi_1 = 1$ is not the critical line when $\rho < 1$, since $\chi_2 < 1$. For learning rates of each network, we consider logarithmically spaced in steps $10^{1}$. To search the optimal learning rate, we select a threshold accuracy of  $p=0.25$ and measure the first step $\tau$ when performance exceeds $p$. We show the steps $\tau$ as a function of learning rate $\eta$ on the networks of dropout rate $\rho = 1.0, 0.99$, and 0.98 in Figure \ref{fig:speed}.

We find that for networks without dropout, there is a universal scaling $\tau = f_1(\eta L)$ between the steps and learning rate, where $f_1$ is a scaling function, as shown in Figure \ref{fig:speed}(a). Note that it is different to the result that $\tau/\sqrt{L} = f’_1(\eta L)$ in \cite{pennington2017resurrecting} where they use the standard CIFAR10 dataset augmented with random flips, crops, and so on. The difference may be caused by the pretreatment of the dataset in \cite{pennington2017resurrecting}. Besides, we study the networks with $\rho = 0.99 $ and $\rho = 0.98$, and find that the scaling $\tau = f_2(\eta)$ can be kept under a limited length $L=120$ for $\rho=0.99$ and $L=55$ for $\rho=0.98$, as shown in Figure \ref{fig:speed}(b) and (c) respectively.

\subsection{Trainable length}

Now we study the problem of trainable length. We consider random networks of depth $10 \le L \le 250$, and $1 \le \sigma^2_w \le 4 $ with $\sigma^2_b = 0.05$. We train these networks using Stochastic Gradient Descent (SGD) and  RMSProp on MNIST and CIFAR10 with Gaussian and Orthogonal weights, which can be seen as another variant of weight initialization in the mean field theory \cite{pennington2017resurrecting}. We perform four experiments on the network without dropout ($\rho=1$) with different datasets, optimizer, and learning rate to conduct a comprehensive study, and plot the results in Figure \ref{fig:heatmap}(a)-(d). Besides, four experiments are conducted on the dropout networks ($\rho<1$), and results are shown in Figure \ref{fig:heatmap}(e)-(h). We color in bright yellow the training accuracy that networks achieved as a function of $\sigma^2_w$ and $L$ for different dropout rates. From the heatmap, we can observe a boundary in which accuracy began to drop. We noticed that there are two boundaries, left and right. In order to show its relationship with $\xi_1$ and $\xi_2$, we superimpose them onto the heatmap.

In figure \ref{fig:heatmap}(a), we use the same learning rate and optimizer as those in Figure 5(a)-(c) of \cite{schoenholz2016deep}. We use a learning rate of $10^{-3}$ for SGD when $L \le 200$, and $10^{-4}$ for larger $L$. From the plot, we find the $6\xi_2$ underestimates the scope of train-ability in the $\sigma^2_w$-$L$ plane, while $12\xi_1$ is more compatible with the experimental result. We note the phenomenon that $6\xi_2$ underestimates the scope of train-ability also happened in Figure 5(b)(c) of \cite{schoenholz2016deep}. In figure \ref{fig:heatmap}(b), we adopt the same learning rate and optimizer as those in Figure 5(d) of \cite{schoenholz2016deep}, where we use a learning of $10^{-5}$ and RMSProp optimizer. Here, the only difference is that we use 1000 training steps instead of 300 training steps in \cite{schoenholz2016deep}. According to the simulation result, $12\xi_1$ (solid line) and $\xi_2$ (dashed line) are identical on the left boundary, while they differ on the right side. We make a comparison between $12\xi_1$ and $12\xi_2$, and find that $12\xi_1$ has a much better argument with the trainable length while $12 \xi_2$ overrates the trainable length on the right side. 

Based on the analysis of Figure \ref{fig:heatmap}(a)(b), we may conclude that $12\xi_1$ can be used to measure the maximum trainable length of the network without dropout. We further reinforce this conclusion by performing experiments on different learning rates, weight initialization, and datasets. In figure(c), we use orthogonal weight initialization. In figure(d), we perform experiment on CIFAR10 dataset and adopt a learning rate of $\eta = c/L$,  where $c$ is constant. These learning rates were selected for the reason that each learning rate can lead to the fast step to a certain test accuracy at $\chi_1= 1 $, as shown in Figure \ref{fig:speed}. In a word, we attribute the maximum trainable length to $L \le \min \{12\xi_1, 12\xi_2 \} = 12\xi_1$, where the relation $\xi_1 \le \xi_2$ holds on the network without dropout.

Furthermore, we consider the dropout case in Figure \ref{fig:heatmap}(e)-(h). We have studied three different dropout rate: $\rho=0.99$ (Figure \ref{fig:heatmap}(e)), $\rho=0.98$ (Figure \ref{fig:heatmap}(f)(g)), and $\rho=0.95$ (Figure \ref{fig:heatmap}(h)). We find that both $\xi_1$ and $\xi_2$ have connections to the trainable length: the networks appear to be trainable when $L \le \min \{12\xi_1, 12\xi_2 \}$. Networks on the left side are influenced by $12 \xi_2$ while they are constrained by the $12 \xi_1$ on the right size. Note that the formula $L \le \min \{12\xi_1, 12\xi_2 \}$ is valid in the no dropout case as discussed above. To conclude, we show an improved relationship between maximum trainable length and length scale $\xi_1$ and $\xi_2$ than \cite{schoenholz2016deep}. This conclusion that both $\xi_1$ and $\xi_2$ have connections to the trainable length instead of only $\xi_2$ \cite{schoenholz2016deep} is more compatible with the theoretical results.

\section{Discussion}

In this paper, we have investigated the dropout networks by calculating its statistical metrics of gradient during the backpropagation at initialization and conjecture that both gradients metric with a single input and a pair of inputs are governed by the same quantity $\chi_1$. We further investigate the relationship between variance and mean of statistical metrics empirically and find an emergence of universality. Our finding of a universal relationship between variance and mean of statistical metrics of gradient backpropagation suggests a deeper mechanism behind it. This mechanism may be comprehended better by studying more different network structures such as Resnet. Finally, for networks with or without dropout, we attribute the maximum trainable length to the formula $L \le \min \{12\xi_1, 12\xi_2 \}$, which is novel and important.

\bibliographystyle{unsrt}
\bibliography{main}

\newpage
\begin{center}
\textbf{\Large {\LARGE Supplemental Material}}
\end{center}

\setcounter{equation}{0}
\setcounter{figure}{0}
\setcounter{table}{0}
\setcounter{section}{0}
\makeatletter
\renewcommand{\theequation}{S\arabic{equation}}
\renewcommand{\thefigure}{S\arabic{figure}}
\renewcommand{\bibnumfmt}[1]{[S#1]}

\section{Theoretical derivation of gradient metrics at initialization}

\subsection{ Derivation of $\widetilde{q}_{aa}^{l}$ on linear dropout networks with a single input}

(1)The $L^{th}$ layer:
\[\delta_{i,a}^{L} = \frac{\partial E_{a}}{\partial z_{i,a}^{L}} =\frac{\partial E_{a}}{\partial y_{i,a}^{L}}=2y_{i,a}^{L}=2z_{i,a}^{L}\]
\[\widetilde{q}_{aa}^{L} = E[(\delta_{i,a}^{L})^{2}] = 4E[(z_{i,a}^{L})^{2}]=4 q_{aa}^{*}\]

(2)The $(L-1)^{th}$ layer:
\[z_{j,a}^{L}=\frac{1}{\rho}\sum_{k}W_{jk}^{L}p_{k,a}^{L}y_{k,a}^{L-1}+b_{j}^{L}, \qquad y_{k,a}^{L-1}=\phi_(z_{k,a}^{L-1})=z_{k,a}^{L-1}\]
\[\delta_{i,a}^{L-1} = \frac{\partial E_{a}}{\partial z_{a}^{L}}\frac{\partial z_{a}^{L}}{\partial z_{i,a}^{L-1}}=\sum_{j}\delta_{j,a}^{L}\frac{\partial z_{j,a}^{L} }{\partial z_{i,a}^{L-1}}=\sum_{j}\delta_{j,a}^{L} \frac{p_{i,a}^{L}}{\rho}W_{ji}^{L}=\sum_{j}2z_{j,a}^{L}\frac{p_{i,a}^{L}}{\rho}W_{ji}^{L}\]
\begin{equation}\label{1}
  \begin{split}
    \widetilde{q}_{aa}^{L-1} & = 4E\Big[\sum_{j}\sum_{j'}z_{j,a}^{L}z_{j',a}^{L}\frac{(p_{i,a}^{L})^{2}}{\rho^{2}}W_{ji}^{L}W_{j'i}^{L}\Big]\\
      & =4E\Big[\sum_{j}\sum_{j'}\sum_{k}\sum_{k'}  (\frac{p_{k,a}^{L}}{\rho}\frac{p_{k',a}^{L}}{\rho}W_{jk}^{L} W_{j'k'}^{L}z_{k,a}^{L-1}z_{k',a}^{L-1}+b_{j}^{L}b_{j'}^{L})\frac{(p_{i,a}^{L})^{2}}{\rho^{2}}W_{ji}^{L}W_{j'i}^{L}\Big]\\
      & =4E\Big[\sum_{j=j'}\sum_{k=k'}\frac{(p_{k,a}^{L})^{2}}{\rho^{2}}\frac{(p_{i,a}^{L})^{2}}{\rho^{2}}(W_{jk}^{L}W_{ji}^{L})^{2}(z_{k,a}^{L-1})^{2}+\sum_{j\neq j'}\sum_{k=k'=i}\frac{(p_{i,a}^{L})^{4}}{\rho^{4}}(W_{ji}^{L}W_{j'i}^{L})^{2}(z_{i,a}^{L-1})^{2}\\
      &+ \sum_{j=j'}(b_{j}^{L})^{2}\frac{(p_{i,a}^{L})^{2}}{\rho^{2}}(W_{ji}^{L})^{2}\Big]\\
      & \approx 4\Big[\frac{1}{\rho^{2}}(\sigma_{\omega}^{2})^{2} q_{aa}^{*}+\frac{1}{\rho^{3}}(\sigma_{\omega}^{2})^{2} q_{aa}^{*}+\sigma_{b}^{2}\sigma_{\omega}^{2}\frac{1}{\rho}\Big]
  \end{split}
\end{equation}
Since,
 \[ q_{aa}^{*}=\frac{\sigma_{\omega}^{2}}{\rho}q_{aa}^{*}+\sigma_{b}^{2}\]
We rewrite Eq (\ref{1}) as:
\begin{equation}\label{2}
  \begin{split}
    \widetilde{q}_{aa}^{L-1} &=4\Big[\frac{1}{\rho^{2}}(\sigma_{\omega}^{2})^{2} q_{aa}^{*}+\frac{1}{\rho^{3}}(\sigma_{\omega}^{2})^{2} q_{aa}^{*}+\sigma_{b}^{2}\sigma_{\omega}^{2}\frac{1}{\rho}\Big]\\
      & =4\Big[\frac{\sigma_{\omega}^{2}}{\rho} q_{aa}^{*}+\frac{1}{\rho^{3}}(\sigma_{\omega}^{2})^{2} q_{aa}^{*}\Big]\\
      &=4\frac{ q_{aa}^{*}}{\rho}\frac{\sigma_{\omega}^{2}}{\rho}\Big[\rho+\frac{\sigma_{\omega}^{2}}{\rho}\Big]
  \end{split}
\end{equation}

(3)The $(L-2)^{th}$ layer:
\[\delta_{i,a}^{L-2} = \sum_{j}\delta_{j,a}^{L-1} \frac{p_{i,a}^{L-1}}{\rho}W_{ji}^{L-1}=\sum_{j}\sum_{k}2z_{k,a}^{L}\frac{p_{j,a}^{L}}{\rho}W_{kj}^{L}\frac{p_{i,a}^{L-1}}{\rho}W_{ji}^{L-1} 
=2\sum_{j}\sum_{k}z_{k,a}^{L}\frac{p_{j,a}^{L}}{\rho}\frac{p_{i,a}^{L-1}}{\rho}W_{kj}^{L}W_{ji}^{L-1}\]
\begin{equation}\label{3}
  \begin{split}
    \widetilde{q}_{aa}^{L-2} & =4E\Big[\sum_{j}\sum_{j'}\sum_{k}\sum_{k'}z_{k,a}^{L}z_{k',a}^{L}\frac{p_{j,a}^{L}p_{j',a}^{L}}{\rho^{2}}\frac{(p_{i,a}^{L-1})^{2}}{\rho^{2}}W_{kj}^{L}W_{k'j'}^{L}W_{ji}^{L-1}W_{j'i}^{L-1}\Big] \\
      & = 4E\Big[\sum_{j}\sum_{j'}\sum_{k}\sum_{k'}\sum_{m}\sum_{m'}(z_{m,a}^{L-1}z_{m',a}^{L-1}\frac{p_{m,a}^{L}p_{m',a}^{L}}{\rho^{2}}W_{km}^{L}W_{k'm'}^{L}+b_{k}^{L}b_{k'}^{L})\frac{p_{j,a}^{L}p_{j',a}^{L}}{\rho^{2}}\frac{(p_{i,a}^{L-1})^{2}}{\rho^{2}}W_{kj}^{L}W_{k'j'}^{L}W_{ji}^{L-1}W_{j'i}^{L-1}\Big]\\
      & =
      4E\Big[\sum_{j}\sum_{j'}\sum_{k}\sum_{k'}\sum_{m}\sum_{m'}z_{m,a}^{L-1}z_{m',a}^{L-1}\frac{p_{m,a}^{L}p_{m',a}^{L}p_{j,a}^{L}p_{j',a}^{L}}{\rho^{4}}\frac{(p_{i,a}^{L-1})^{2}}{\rho^{2}}W_{km}^{L}W_{k'm'}^{L}W_{kj}^{L}W_{k'j'}^{L}W_{ji}^{L-1}W_{j'i}^{L-1}\\
      & +\sum_{j}\sum_{j'}\sum_{k}\sum_{k'}b_{k}^{L}b_{k'}^{L}\frac{p_{j,a}^{L}p_{j',a}^{L}}{\rho^{2}}\frac{(p_{i,a}^{L-1})^{2}}{\rho^{2}}W_{kj}^{L}W_{k'j'}^{L}W_{ji}^{L-1}W_{j'i}^{L-1}\Big]\\
      &=4E\Big[\sum_{j,j',k,k' \atop m,m',n,n'}(\frac{p_{n,a}^{L-1}p_{n',a}^{L-1}}{\rho^{2}}W_{mn}^{L-1}W_{m'n'}^{L-1}z_{n,a}^{L-2}z_{n',a}^{L-2}+b_{m}^{L-1}b_{m'}^{L-1})\frac{p_{m,a}^{L}p_{m',a}^{L}p_{j,a}^{L}p_{j',a}^{L}}{\rho^{4}}\frac{(p_{i,a}^{L-1})^{2}}{\rho^{2}} \\
      & W_{km}^{L}W_{k'm'}^{L}W_{kj}^{L}W_{k'j'}^{L}W_{ji}^{L-1}W_{j'i}^{L-1}\\
      &+\sum_{j,j',k,k'}b_{k}^{L}b_{k'}^{L}\frac{p_{j,a}^{L}p_{j',a}^{L}}{\rho^{2}}\frac{(p_{i,a}^{L-1})^{2}}{\rho^{2}}W_{kj}^{L}W_{k'j'}^{L}W_{ji}^{L-1}W_{j'i}^{L-1}\Big]\\
      &=4E\Big[\sum_{j,j',k,k' \atop m,m',n,n'}z_{n,a}^{L-2}z_{n',a}^{L-2}\frac{p_{m,a}^{L}p_{m',a}^{L}p_{j,a}^{L}p_{j',a}^{L}}{\rho^{4}}\frac{p_{n,a}^{L-1}p_{n',a}^{L-1}(p_{i,a}^{L-1})^{2}}{\rho^{4}}W_{km}^{L}W_{k'm'}^{L}W_{kj}^{L}W_{k'j'}^{L}W_{ji}^{L-1}W_{j'i}^{L-1}W_{mn}^{L-1}W_{m'n'}^{L-1}\\
  &+\sum_{j,j',k,k',m,m'}b_{m}^{L-1}b_{m'}^{L-1}\frac{p_{m,a}^{L}p_{m',a}^{L}p_{j,a}^{L}p_{j',a}^{L}}{\rho^{4}}\frac{(p_{i,a}^{L-1})^{2}}{\rho^{2}}W_{km}^{L}W_{k'm'}^{L}W_{kj}^{L}W_{k'j'}^{L}W_{ji}^{L-1}W_{j'i}^{L-1}\\
      &+\sum_{j,j',k,k'}b_{k}^{L}b_{k'}^{L}\frac{p_{j,a}^{L}p_{j',a}^{L}}{\rho^{2}}\frac{(p_{i,a}^{L-1})^{2}}{\rho^{2}}W_{kj}^{L}W_{k'j'}^{L}W_{ji}^{L-1}W_{j'i}^{L-1}\Big]\\
      &=4E\Big[ {\rm I+II+III }\Big]
  \end{split}
\end{equation}
There are three parts in Eq (\ref{3}),  we denote them as I, II, III and compute them one by one,
\begin{equation}\label{4}
  \begin{split}
  E[{\rm I}]& =E\Big[\sum_{j,j',k,k' \atop m,m',n,n'}z_{n,a}^{L-2}z_{n',a}^{L-2}\frac{p_{m,a}^{L}p_{m',a}^{L}p_{j,a}^{L}p_{j',a}^{L}}{\rho^{4}}\frac{p_{n,a}^{L-1}p_{n',a}^{L-1}(p_{i,a}^{L-1})^{2}}{\rho^{4}}W_{km}^{L}W_{k'm'}^{L}W_{kj}^{L}W_{k'j'}^{L}W_{ji}^{L-1}W_{j'i}^{L-1}W_{mn}^{L-1}W_{m'n'}^{L-1}\Big]\\
  &=E\Big[(\sum_{\mbox{\tiny$\begin{array}{c}
  n=n'=i\\
  k\neq k'\\
  m=j\\
  m'=j'
  \end{array}$}}+\sum_{\mbox{\tiny$\begin{array}{c}
  n=n'\neq i\\
  k=k'\\
  m=m'\\
  j=j'
  \end{array}$}}+\sum_{\mbox{\tiny$\begin{array}{c}
  n=n'\neq i\\
  k\neq k'\\
  m=m'=j=j'
  \end{array}$}})z_{n,a}^{L-2}z_{n',a}^{L-2}\frac{p_{m,a}^{L}p_{m',a}^{L}p_{j,a}^{L}p_{j',a}^{L}}{\rho^{4}}\frac{p_{n,a}^{L-1}p_{n',a}^{L-1}(p_{i,a}^{L-1})^{2}}{\rho^{4}}\\
  &W_{km}^{L}W_{k'm'}^{L}W_{kj}^{L}W_{k'j'}^{L}W_{ji}^{L-1}W_{j'i}^{L-1}W_{mn}^{L-1}W_{m'n'}^{L-1}\Big]\\
  &=E\Big[\sum_{n=n'=i, k\neq k'\atop m=j, m'=j'}(z_{i,a}^{L-2})^{2}\frac{(p_{j,a}^{L})^{2}(p_{j',a}^{L})^{2}}{\rho^{4}}\frac{(p_{i,a}^{L-1})^{4}}{\rho^{4}}(W_{kj}^{L}W_{k'j'})^{2}(W_{ji}^{L-1}W_{j'i}^{L-1})^{2}\\
  &+\sum_{n=n'\neq i, k=k'\atop m=m',j=j'}(z_{n,a}^{L-2})^{2}\frac{(p_{m,a}^{L})^{2}(p_{j,a}^{L})^{2}}{\rho^{4}}\frac{(p_{n,a}^{L-1})^{2}(p_{i,a}^{L-1})^{2}}{\rho^{4}}(W_{km}^{L}W_{kj}^{L})^{2}(W_{ji}^{L-1}W_{mn}^{L-1})^{2}\\
  &+\sum_{n=n'\neq i, k\neq k'\atop m=m'=j=j'}(z_{n,a}^{L-2})^{2}\frac{(p_{j,a}^{L})^{4}}{\rho^{4}}\frac{(p_{n,a}^{L-1})^{2}(p_{i,a}^{L-1})^{2}}{\rho^{4}}(W_{kj}^{L}W_{k'j}^{L})^{2}(W_{ji}^{L-1}W_{jn}^{L-1})^{2}\Big]\\
  & \approx q_{aa}^{L-2}\frac{1}{\rho^{5}}(\sigma_{\omega}^{2})^{4}+q_{aa}^{L-2}\frac{1}{\rho^{4}}(\sigma_{\omega}^{2})^{4}+q_{aa}^{L-2}\frac{1}{\rho^{5}}(\sigma_{\omega}^{2})^{4}\\
  &=q_{aa}^{L-2}\Big[\frac{(\sigma_{\omega}^{2})^{4}}{\rho^{5}}+\frac{(\sigma_{\omega}^{2})^{4}}{\rho^{4}}+\frac{(\sigma_{\omega}^{2})^{4}}{\rho^{5}}\Big]
  \end{split}
\end{equation}

\begin{equation}\label{5}
  \begin{split}
  E[{\rm II}]& =E\Big[\sum_{j,j',k,k',m,m'}b_{m}^{L-1}b_{m'}^{L-1}\frac{p_{m,a}^{L}p_{m',a}^{L}p_{j,a}^{L}p_{j',a}^{L}}{\rho^{4}}\frac{(p_{i,a}^{L-1})^{2}}{\rho^{2}}W_{km}^{L}W_{k'm'}^{L}W_{kj}^{L}W_{k'j'}^{L}W_{ji}^{L-1}W_{j'i}^{L-1}\Big]\\
  &=E\Big[(\sum_{\mbox{\tiny$\begin{array}{c}
  k\neq k'\\
  m=j=m'=j'
  \end{array}$}}+\sum_{\mbox{\tiny$\begin{array}{c}
  k=k'\\
  m=m'\\
  j=j'
  \end{array}$}})b_{m}^{L-1}b_{m'}^{L-1}\frac{p_{m,a}^{L}p_{m',a}^{L}p_{j,a}^{L}p_{j',a}^{L}}{\rho^{4}}\frac{(p_{i,a}^{L-1})^{2}}{\rho^{2}}W_{km}^{L}W_{k'm'}^{L}W_{kj}^{L}W_{k'j'}^{L}W_{ji}^{L-1}W_{j'i}^{L-1}\Big]\\
  &=E\Big[\sum_{k\neq k'\atop m=j=m'=j'}(b_{m}^{L-1})^{2}\frac{(p_{j,a}^{L})^{4}}{\rho^{4}}\frac{(p_{i,a}^{L-1})^{2}}{\rho^{2}}(W_{kj}^{L}W_{k'j'}^{L})^{2}(W_{ji}^{L-1})^{2}\Big]\\
  &+\sum_{k=k',m=m',j=j'}(b_{m}^{L-1})^{2}\frac{(p_{m,a}^{L})^{2}(p_{j,a}^{L})^{2}}{\rho^{4}}\frac{(p_{i,a}^{L-1})^{2}}{\rho^{2}}(W_{km}^{L}W_{kj}^{L})^{2}(W_{ji}^{L-1})^{2}\Big]\\
  &\approx \sigma_{b}^{2}\frac{(\sigma_{\omega}^{2})^{3}}{\rho^{4}}+\sigma_{b}^{2}\frac{(\sigma_{\omega}^{2})^{3}}{\rho^{3}}
  \end{split}
\end{equation}

\begin{equation}\label{6}
  \begin{split}
  E[{\rm III}]& = E\Big[\sum_{j,j',k,k'}b_{k}^{L}b_{k'}^{L}\frac{p_{j,a}^{L}p_{j',a}^{L}}{\rho^{2}}\frac{(p_{i,a}^{L-1})^{2}}{\rho^{2}}W_{kj}^{L}W_{k'j'}^{L}W_{ji}^{L-1}W_{j'i}^{L-1}\Big]\\
  &=E\Big[\sum_{j=j',k=k'}(b_{k}^{L})^{2}\frac{(p_{j,a}^{L})^{2}}{\rho^{2}}\frac{(p_{i,a}^{L-1})^{2}}{\rho^{2}}(W_{kj}^{L})^{2}(W_{ji}^{L-1})^{2}\Big]\\
  &= \sigma_{b}^{2}\frac{(\sigma_{\omega}^{2})^{2}}{\rho^{2}}
  \end{split}
\end{equation}

Finally, we have,
\begin{equation}\label{7}
  \begin{split}
  \widetilde{q}_{aa}^{L-2}& =4E\Big[{\rm I+II+III } \Big]\\
  &= 4\Big[q_{aa}^{*}\Big((\frac{(\sigma_{\omega}^{2})^{4}}{\rho^{5}}+\frac{(\sigma_{\omega}^{2})^{4}}{\rho^{4}}+\frac{(\sigma_{\omega}^{2})^{4}}{\rho^{5}}\Big)
  +\sigma_{b}^{2}\frac{(\sigma_{\omega}^{2})^{3}}{\rho^{4}}+\sigma_{b}^{2}\frac{(\sigma_{\omega}^{2})^{3}}{\rho^{3}}
  +\sigma_{b}^{2}\frac{(\sigma_{\omega}^{2})^{2}}{\rho^{2}}\Big]\\
  &=4\Big[q_{aa}^{*}\frac{(\sigma_{\omega}^{2})^{4}}{\rho^{5}}+q_{aa}^{*}\frac{(\sigma_{\omega}^{2})^{3}}{\rho^{4}}
  +q_{aa}^{*}\frac{(\sigma_{\omega}^{2})^{3}}{\rho^{3}}+\sigma_{b}^{2}\frac{(\sigma_{\omega}^{2})^{2}}{\rho^{2}}\Big]\\
  &=4\Big[q_{aa}^{*}\frac{(\sigma_{\omega}^{2})^{4}}{\rho^{5}}+q_{aa}^{*}\frac{(\sigma_{\omega}^{2})^{3}}{\rho^{4}}+
  q_{aa}^{*}\frac{(\sigma_{\omega}^{2})^{2}}{\rho^{2}}\Big]\\
  &=4\frac{q_{aa}^{*}}{\rho}(\frac{\sigma_{\omega}^{2}}{\rho})^{2}\Big[\rho+\frac{\sigma_{\omega}^{2}}{\rho}+(\frac{\sigma_{\omega}^{2}}{\rho})^{2}\Big]
  \end{split}
\end{equation}

To summarize, we list the results for $\widetilde{q}^L_{aa}$, $\widetilde{q}^{L-1}_{aa}$, and $\widetilde{q}^{L-2}_{aa}$:
\begin{eqnarray}
  \widetilde{q}_{aa}^{L} &=& 4 q_{aa}^{*} \\
  \widetilde{q}_{aa}^{L-1} &=& 4\frac{q_{aa}^{*}}{\rho}\frac{\sigma_{\omega}^{2}}{\rho}\Big[\rho+\frac{\sigma_{\omega}^{2}}{\rho}\Big]\\
  \widetilde{q}_{aa}^{L-2} &=& 4\frac{q_{aa}^{*}}{\rho}\Big(\frac{\sigma_{\omega}^{2}}{\rho}\Big)^{2}\Big[\rho+\frac{\sigma_{\omega}^{2}}{\rho}+(\frac{\sigma_{\omega}^{2}}{\rho})^{2}\Big]
\end{eqnarray}
Using mathematical induction method we draw the conclusion that,
\begin{equation}
 \widetilde{q}_{aa}^{l}=4\frac{q_{aa}^{*}}{\rho}\Big(\frac{\sigma_{\omega}^{2}}{\rho}\Big)^{L-l}\Big[\rho+\sum_{j=1}^{L-l}\Big(\frac{\sigma_{\omega}^{2}}{\rho}\Big)^{j}\Big] 
\end{equation}
Using the relation,
\begin{equation}
    \frac{\partial E}{\partial W^{l}_{ij}} = \frac{ p^l_j}{\rho}  \phi(z^{l-1}_j) \delta^l_i,
  \end{equation}
we obtain the final result,
\begin{equation}
 g_{aa}^{l}=4(\frac{q_{aa}^{*}}{\rho})^2 \Big(\frac{\sigma_{\omega}^{2}}{\rho}\Big)^{L-l}\Big[\rho+\sum_{j=1}^{L-l}\Big(\frac{\sigma_{\omega}^{2}}{\rho}\Big)^{j}\Big]. 
\end{equation}

\newpage

\subsection{Derivation of $\widetilde{q}_{ab}^{l}$ on linear dropout networks with a pair of inputs}
(1)The $L^{th}$ layer:
\[\delta_{i,a}^{L} =2z_{i,a}^{L}\]
\[\widetilde{q}_{ab}^{L} = E[(\delta_{i,a}^{L}\delta_{i,b}^{L})]=4E[(z_{i,a}^{L}z_{i,b}^{L})]=4 q_{ab}^{*}\]
(2)The $(L-1)^{th}$ layer:
\[z_{j,a}^{L}=\frac{1}{\rho}\sum_{k}W_{jk}^{L}p_{k,a}^{L}y_{k,a}^{L-1}+b_{j}^{L}, \qquad y_{k,a}^{L-1}=\phi_(z_{k,a}^{L-1})=z_{k,a}^{L-1}\]
\[z_{j',b}^{L}=\frac{1}{\rho}\sum_{k'}W_{j'k'}^{L}p_{k',b}^{L}y_{k',b}^{L-1}+b_{j}^{L}, \qquad y_{k',b}^{L-1}=\phi_(z_{k',b}^{L-1})=z_{k',b}^{L-1}\]
\[\delta_{i,a}^{L-1} =\sum_{j}2z_{j,a}^{L}\frac{p_{i,a}^{L}}{\rho}W_{ji}^{L}\] \qquad  \[\delta_{i,b}^{L-1} =\sum_{j'}2z_{j',b}^{L}\frac{p_{i,b}^{L}}{\rho}W_{j'i}^{L}\]

\begin{equation}\label{12}
  \begin{split}
    \widetilde{q}_{ab}^{L-1} & = 4E\Big[\sum_{j}\sum_{j'}z_{j,a}^{L}z_{j',b}^{L}\frac{p_{i,a}^{L}p_{i,b}^{L}}{\rho^{2}}W_{ji}^{L}W_{j'i}^{L}\Big]\\
      & =4E\Big[\sum_{j}\sum_{j'}\sum_{k}\sum_{k'}  (\frac{p_{k,a}^{L}}{\rho}\frac{p_{k',b}^{L}}{\rho}W_{jk}^{L} W_{j'k'}^{L}z_{k,a}^{L-1}z_{k',b}^{L-1}+b_{j}^{L}b_{j'}^{L})\frac{p_{i,a}^{L}p_{i,b}^{L}}{\rho^{2}}W_{ji}^{L}W_{j'i}^{L}\Big]\\
      & =4E\Big[\sum_{j=j'}\sum_{k=k'}z_{k,a}^{L-1}z_{k,b}^{L-1}\frac{p_{k,a}^{L}p_{k,b}^{L}p_{i,a}^{L}p_{i,b}^{L}}{\rho^{4}}(W_{jk}^{L}W_{ji}^{L})^{2}+\sum_{j\neq j'}\sum_{k=k'=i}z_{i,a}^{L-1}z_{i,b}^{L-1}\frac{(p_{i,a}^{L})^{2}(p_{i,b}^{L})^{2}}{\rho^{4}}(W_{ji}^{L}W_{j'i}^{L})^{2}\\
      &+ \sum_{j=j'}(b_{j}^{L})^{2}\frac{p_{i,a}^{L}p_{i,b}^{L}}{\rho^{2}}(W_{ji}^{L})^{2}\Big]\\
      & \approx 4\Big[(\sigma_{\omega}^{2})^{2}q_{ab}^{*}+\frac{1}{\rho^{2}}(\sigma_{\omega}^{2})^{2}q_{ab}^{*}+\sigma_{b}^{2}\sigma_{\omega}^{2}\Big]
  \end{split}
\end{equation}
Here, we have
\[ q_{ab}^{*}=\sigma_{\omega}^{2} q_{ab}^{*}+\sigma_{b}^{2}\]
We rewrite Eq (\ref{12}) as:
\begin{equation}\label{13}
  \begin{split}
    \widetilde{q}_{ab}^{L-1} &=4\Big[\frac{1}{\rho^{2}}(\sigma_{\omega}^{2})^{2} q_{ab}^{*}+(\sigma_{\omega}^{2})^{2} q_{ab}^{*}+\sigma_{b}^{2}\sigma_{\omega}^{2}\Big]\\
      & =4\Big[\frac{1}{\rho^{2}}(\sigma_{\omega}^{2})^{2} q_{ab}^{*}+\sigma_{\omega}^{2} q_{ab}^{*}\Big]\\
      &=4 q_{ab}^{*}\sigma_{\omega}^{2}\Big[1+\frac{\sigma_{\omega}^{2}}{\rho^{2}}\Big]
  \end{split}
\end{equation}
(3)The $(L-2)^{th}$ layer:
\[\delta_{i,a}^{L-2} = \sum_{j}\delta_{j,a}^{L-1} \frac{p_{i,a}^{L-1}}{\rho}W_{ji}^{L-1}=\sum_{j}\sum_{k}2z_{k,a}^{L}\frac{p_{j,a}^{L}}{\rho}W_{kj}^{L}\frac{p_{i,a}^{L-1}}{\rho}W_{ji}^{L-1}
=2\sum_{j}\sum_{k}z_{k,a}^{L}\frac{p_{j,a}^{L}}{\rho}\frac{p_{i,a}^{L-1}}{\rho}W_{kj}^{L}W_{ji}^{L-1}\]
\[\delta_{i,b}^{L-2} = \sum_{j'}\delta_{j',b}^{L-1} \frac{p_{i,b}^{L-1}}{\rho}W_{j'i}^{L-1}=\sum_{j'}\sum_{k'}2z_{k',b}^{L}\frac{p_{j',b}^{L}}{\rho}W_{k'j'}^{L}\frac{p_{i,b}^{L-1}}{\rho}W_{j'i}^{L-1}
=2\sum_{j'}\sum_{k'}z_{k',b}^{L}\frac{p_{j',b}^{L}}{\rho}\frac{p_{i,b}^{L-1}}{\rho}W_{k'j'}^{L}W_{j'i}^{L-1}\]
\begin{equation}\label{14}
  \begin{split}
    \widetilde{q}_{ab}^{L-2} & =4E\Big[\sum_{j}\sum_{j'}\sum_{k}\sum_{k'}z_{k,a}^{L}z_{k',b}^{L}\frac{p_{j,a}^{L}p_{j',b}^{L}}{\rho^{2}}\frac{p_{i,a}^{L-1}p_{i,b}^{L-1}}{\rho^{2}}W_{kj}^{L}W_{k'j'}^{L}W_{ji}^{L-1}W_{j'i}^{L-1}\Big] \\
      & = 4E\Big[\sum_{j}\sum_{j'}\sum_{k}\sum_{k'}\sum_{m}\sum_{m'}(z_{m,a}^{L-1}z_{m',b}^{L-1}\frac{p_{m,a}^{L}p_{m',b}^{L}}{\rho^{2}}W_{km}^{L}W_{k'm'}^{L}+b_{k}^{L}b_{k'}^{L})\frac{p_{j,a}^{L}p_{j',b}^{L}}{\rho^{2}}\frac{p_{i,a}^{L-1}p_{i,b}^{L-1}}{\rho^{2}}W_{kj}^{L}W_{k'j'}^{L}W_{ji}^{L-1}W_{j'i}^{L-1}\Big]\\
      & =
      4E\Big[\sum_{j}\sum_{j'}\sum_{k}\sum_{k'}\sum_{m}\sum_{m'}z_{m,a}^{L-1}z_{m',b}^{L-1}\frac{p_{m,a}^{L}p_{m',b}^{L}p_{j,a}^{L}p_{j',b}^{L}}{\rho^{4}}\frac{p_{i,a}^{L-1}p_{i,b}^{L-1}}{\rho^{2}}W_{km}^{L}W_{k'm'}^{L}W_{kj}^{L}W_{k'j'}^{L}W_{ji}^{L-1}W_{j'i}^{L-1}\\
      & +\sum_{j}\sum_{j'}\sum_{k}\sum_{k'}b_{k}^{L}b_{k'}^{L}\frac{p_{j,a}^{L}p_{j',b}^{L}}{\rho^{2}}\frac{p_{i,a}^{L-1}p_{i,b}^{L-1}}{\rho^{2}}W_{kj}^{L}W_{k'j'}^{L}W_{ji}^{L-1}W_{j'i}^{L-1}\Big]\\
      &=4E\Big[\sum_{j,j',k,k' \atop m,m',n,n'}(\frac{p_{n,a}^{L-1}p_{n',b}^{L-1}}{\rho^{2}}W_{mn}^{L-1}W_{m'n'}^{L-1}z_{n,a}^{L-2}z_{n',b}^{L-2}+b_{m}^{L-1}b_{m'}^{L-1})\frac{p_{m,a}^{L}p_{m',b}^{L}p_{j,a}^{L}p_{j',b}^{L}}{\rho^{4}}\frac{p_{i,a}^{L-1}p_{i,b}^{L-1}}{\rho^{2}}\\
      &W_{km}^{L}W_{k'm'}^{L}W_{kj}^{L}W_{k'j'}^{L}W_{ji}^{L-1}W_{j'i}^{L-1}+\sum_{j,j',k,k'}b_{k}^{L}b_{k'}^{L}\frac{p_{j,a}^{L}p_{j',b}^{L}}{\rho^{2}}\frac{p_{i,a}^{L-1}p_{i,b}^{L-1}}{\rho^{2}}W_{kj}^{L}W_{k'j'}^{L}W_{ji}^{L-1}W_{j'i}^{L-1}\Big]\\
      &=4E\Big[\sum_{j,j',k,k' \atop m,m',n,n'}z_{n,a}^{L-2}z_{n',b}^{L-2}\frac{p_{m,a}^{L}p_{m',b}^{L}p_{j,a}^{L}p_{j',b}^{L}}{\rho^{4}}\frac{p_{n,a}^{L-1}p_{n',b}^{L-1}p_{i,a}^{L-1}p_{i,b}^{L-1}}{\rho^{4}}W_{km}^{L}W_{k'm'}^{L}W_{kj}^{L}W_{k'j'}^{L}W_{ji}^{L-1}W_{j'i}^{L-1}W_{mn}^{L-1}W_{m'n'}^{L-1}\\
      &+\sum_{j,j',k,k',m,m'}b_{m}^{L-1}b_{m'}^{L-1}\frac{p_{m,a}^{L}p_{m',b}^{L}p_{j,a}^{L}p_{j',b}^{L}}{\rho^{4}}\frac{p_{i,a}^{L-1}p_{i,b}^{L-1}}{\rho^{2}}W_{km}^{L}W_{k'm'}^{L}W_{kj}^{L}W_{k'j'}^{L}W_{ji}^{L-1}W_{j'i}^{L-1}\\
      &+\sum_{j,j',k,k'}b_{k}^{L}b_{k'}^{L}\frac{p_{j,a}^{L}p_{j',b}^{L}}{\rho^{2}}\frac{p_{i,a}^{L-1}p_{i,b}^{L-1}}{\rho^{2}}W_{kj}^{L}W_{k'j'}^{L}W_{ji}^{L-1}W_{j'i}^{L-1}\Big]\\
      &=4E\Big[{\rm I+II+III}\Big]
  \end{split}
\end{equation}

\begin{equation}\label{15}
  \begin{split}
  E[{\rm I}]& =E\Big[\sum_{j,j',k,k' \atop m,m',n,n'}z_{n,a}^{L-2}z_{n',b}^{L-2}\frac{p_{m,a}^{L}p_{m',b}^{L}p_{j,a}^{L}p_{j',b}^{L}}{\rho^{4}}\frac{p_{n,a}^{L-1}p_{n',b}^{L-1}p_{i,a}^{L-1}p_{i,b}^{L-1}}{\rho^{4}}W_{km}^{L}W_{k'm'}^{L}W_{kj}^{L}W_{k'j'}^{L}W_{ji}^{L-1}W_{j'i}^{L-1}W_{mn}^{L-1}W_{m'n'}^{L-1}\Big]\\
  &=E\Big[(\sum_{\mbox{\tiny$\begin{array}{c}
  n=n'=i\\
  k\neq k'\\
  m=j\\
  m'=j'
  \end{array}$}}+\sum_{\mbox{\tiny$\begin{array}{c}
  n=n'\neq i\\
  k=k'\\
  m=m'\\
  j=j'
  \end{array}$}}+\sum_{\mbox{\tiny$\begin{array}{c}
  n=n'\neq i\\
  k\neq k'\\
  m=m'=j=j'
  \end{array}$}})z_{n,a}^{L-2}z_{n',b}^{L-2}\frac{p_{m,a}^{L}p_{m',b}^{L}p_{j,a}^{L}p_{j',b}^{L}}{\rho^{4}}\frac{p_{n,a}^{L-1}p_{n',b}^{L-1}p_{i,a}^{L-1}p_{i,b}^{L-1}}{\rho^{4}}\\
  &W_{km}^{L}W_{k'm'}^{L}W_{kj}^{L}W_{k'j'}^{L}W_{ji}^{L-1}W_{j'i}^{L-1}W_{mn}^{L-1}W_{m'n'}^{L-1}\Big]\\
  &=E\Big[\sum_{n=n'=i, k\neq k'\atop m=j, m'=j'}z_{n,a}^{L-2}z_{n,b}^{L-2}\frac{(p_{j,a}^{L}p_{j',b}^{L})^{2}}{\rho^{4}}\frac{(p_{i,a}^{L-1}p_{i,b}^{L-1})^{2}}{\rho^{4}}(W_{kj}^{L}W_{k'j'})^{2}(W_{ji}^{L-1}W_{j'i}^{L-1})^{2}\\
  &+\sum_{n=n'\neq i, k=k'\atop m=m',j=j'}z_{n,a}^{L-2}z_{n,b}^{L-2}\frac{p_{m,a}^{L}p_{m,b}^{L}p_{j,a}^{L}p_{j,b}^{L}}{\rho^{4}}\frac{p_{n,a}^{L-1}p_{n,b}^{L-1}p_{i,a}^{L-1}p_{i,b}^{L-1}}{\rho^{4}}(W_{km}^{L}W_{kj}^{L})^{2}(W_{ji}^{L-1}W_{mn}^{L-1})^{2}\\
  &+\sum_{n=n'\neq i, k\neq k'\atop m=m'=j=j'}(z_{n,a}^{L-2})^{2}\frac{(p_{j,a}^{L}p_{j,b}^{L})^{2}}{\rho^{4}}\frac{p_{n,a}^{L-1}p_{n,b}^{L-1}p_{i,a}^{L-1}p_{i,b}^{L-1}}{\rho^{4}}(W_{kj}^{L}W_{k'j}^{L})^{2}(W_{ji}^{L-1}W_{jn}^{L-1})^{2}\Big]\\
  &\approx q_{ab}^{\ast}\frac{1}{\rho^{4}}(\sigma_{\omega}^{2})^{4}+q_{ab}^{\ast}(\sigma_{\omega}^{2})^{4}+q_{ab}^{\ast}\frac{1}{\rho^{2}}(\sigma_{\omega}^{2})^{4}\\
  &= q_{ab}^{\ast}\Big[\frac{(\sigma_{\omega}^{2})^{4}}{\rho^{4}}+(\sigma_{\omega}^{2})^{4}+\frac{(\sigma_{\omega}^{2})^{4}}{\rho^{2}}\Big]
  \end{split}
\end{equation}

\begin{equation}\label{16}
  \begin{split}
  E[{\rm II}]& =E\Big[\sum_{j,j',k,k',m,m'}b_{m}^{L-1}b_{m'}^{L-1}\frac{p_{m,a}^{L}p_{m',b}^{L}p_{j,a}^{L}p_{j',b}^{L}}{\rho^{4}}\frac{p_{i,a}^{L-1}p_{i,b}^{L-1}}{\rho^{2}}W_{km}^{L}W_{k'm'}^{L}W_{kj}^{L}W_{k'j'}^{L}W_{ji}^{L-1}W_{j'i}^{L-1}\Big]\\
  &=E\Big[(\sum_{\mbox{\tiny$\begin{array}{c}
  k\neq k'\\
  m=j=m'=j'
  \end{array}$}}+\sum_{\mbox{\tiny$\begin{array}{c}
  k=k'\\
  m=m'\\
  j=j'
  \end{array}$}})b_{m}^{L-1}b_{m'}^{L-1}\frac{p_{m,a}^{L}p_{m',b}^{L}p_{j,a}^{L}p_{j',b}^{L}}{\rho^{4}}\frac{p_{i,a}^{L-1}p_{i,b}^{L-1}}{\rho^{2}}W_{km}^{L}W_{k'm'}^{L}W_{kj}^{L}W_{k'j'}^{L}W_{ji}^{L-1}W_{j'i}^{L-1}\Big]\\
  &=E\Big[\sum_{k\neq k'\atop m=j=m'=j'}(b_{m}^{L-1})^{2}\frac{(p_{j,a}^{L}p_{j,b}^{L})^{2}}{\rho^{4}}\frac{p_{i,a}^{L-1}p_{i,b}^{L-1}}{\rho^{2}}(W_{kj}^{L}W_{k'j'}^{L})^{2}(W_{ji}^{L-1})^{2}\Big]\\
  &+\sum_{k=k',m=m',j=j'}(b_{m}^{L-1})^{2}\frac{p_{m,a}^{L}p_{m,b}^{L}p_{j,a}^{L}p_{j,b}^{L}}{\rho^{4}}\frac{p_{i,a}^{L}p_{i,b}^{L}}{\rho^{2}}(W_{km}^{L}W_{kj}^{L})^{2}(W_{ji}^{L-1})^{2}\Big]\\
  &\approx \sigma_{b}^{2}\frac{(\sigma_{\omega}^{2})^{3}}{\rho^{2}}+\sigma_{b}^{2}(\sigma_{\omega}^{2})^{3}
  \end{split}
\end{equation}

\begin{equation}\label{17}
  \begin{split}
  E[{\rm III}]& =E\Big[\sum_{j,j',k,k'}b_{k}^{L}b_{k'}^{L}\frac{p_{j,a}^{L}p_{j',b}^{L}}{\rho^{2}}\frac{p_{i,a}^{L-1}p_{i,b}^{L-1}}{\rho^{2}}W_{kj}^{L}W_{k'j'}^{L}W_{ji}^{L-1}W_{j'i}^{L-1}\Big]\\
  &=E\Big[\sum_{j=j',k=k'}(b_{k}^{L})^{2}\frac{p_{j,a}^{L}p_{j,b}^{L}}{\rho^{2}}\frac{p_{i,a}^{L-1}p_{i,b}^{L-1}}{\rho^{2}}(W_{kj}^{L})^{2}(W_{ji}^{L-1})^{2}\Big]\\
  &=\sigma_{b}^{2}(\sigma_{\omega}^{2})^{2}
  \end{split}
\end{equation}

Finally, we have,
\begin{equation}\label{18}
  \begin{split}
  \widetilde{q}_{ab}^{L-2}& =4E\Big[{\rm I+II+III} \Big]\\
  &= 4\Big[q_{ab}^{*}\Big(\frac{(\sigma_{\omega}^{2})^{4}}{\rho^{4}}+(\sigma_{\omega}^{2})^{4}+\frac{(\sigma_{\omega}^{2})^{4}}{\rho^{2}}\Big)
  +\sigma_{b}^{2}\frac{(\sigma_{\omega}^{2})^{3}}{\rho^{2}}+\sigma_{b}^{2}(\sigma_{\omega}^{2})^{3}
  +\sigma_{b}^{2}(\sigma_{\omega}^{2})^{2}\Big]\\
  &=4\Big[q_{ab}^{*}\frac{(\sigma_{\omega}^{2})^{4}}{\rho^{4}}+q_{ab}^{*}\frac{(\sigma_{\omega}^{2})^{3}}{\rho^{2}}
  +q_{ab}^{*}(\sigma_{\omega}^{2})^{3}+\sigma_{b}^{2}(\sigma_{\omega}^{2})^{2}\Big]\\
  &=4\Big[q_{ab}^{*}\frac{(\sigma_{\omega}^{2})^{4}}{\rho^{4}}+q_{ab}^{*}\frac{(\sigma_{\omega}^{2})^{3}}{\rho^{2}}+
  q_{ab}^{*}(\sigma_{\omega}^{2})^{2}\Big]\\
  &=4q_{ab}^{*}(\sigma_{\omega}^{2})^{2}\Big[1+\frac{\sigma_{\omega}^{2}}{\rho^{2}}+(\frac{\sigma_{\omega}^{2}}{\rho^{2}})^{2}\Big]
  \end{split}
\end{equation}

To summarize, we list the results for $\widetilde{q}^L_{ab}$, $\widetilde{q}^{L-1}_{ab}$, and $\widetilde{q}^{L-2}_{ab}$:
\begin{eqnarray}
  \widetilde{q}_{ab}^{L}   &=& 4 q_{ab}^{*} \\
  \widetilde{q}_{ab}^{L-1} &=& 4 q_{ab}^{*}\sigma_{\omega}^{2}\Big[1+\frac{\sigma_{\omega}^{2}}{\rho^{2}}\Big]\\
  \widetilde{q}_{ab}^{L-2} &=& 4 q_{ab}^{*}(\sigma_{\omega}^{2})^{2}\Big[1+\frac{\sigma_{\omega}^{2}}{\rho^{2}}+(\frac{\sigma_{\omega}^{2}}{\rho^{2}})^{2}\Big]
\end{eqnarray}
Using mathematical induction method we draw the conclusion that
\begin{equation}
\widetilde{q}_{ab}^{l}=4 q_{ab}^{*}\Big(\sigma_{\omega}^{2}\Big)^{L-l}\Big[1+\sum_{j=1}^{L-l}\Big(\frac{\sigma_{\omega}^{2}}{\rho^{2}}\Big)^{j}\Big]
\end{equation}

Using the relation,
\begin{equation}
    \frac{\partial E}{\partial W^{l}_{ij}} = \frac{ p^l_j}{\rho}  \phi(z^{l-1}_j) \delta^l_i,
  \end{equation}
we obtain the final result,
\begin{equation}
g_{ab}^{l}=4 (q_{ab}^{*})^2 \Big(\sigma_{\omega}^{2}\Big)^{L-l}\Big[1+\sum_{j=1}^{L-l}\Big(\frac{\sigma_{\omega}^{2}}{\rho^{2}}\Big)^{j}\Big].
\end{equation}

\end{document}